\documentclass[10pt,twocolumn,letterpaper]{article}

\usepackage{iccv}
\usepackage{times}
\usepackage{epsfig}
\usepackage{graphicx}
\usepackage{amsmath}
\usepackage{amssymb}

\usepackage{booktabs}
\usepackage{multirow}
\usepackage{caption}
\usepackage{float}
\usepackage{url}
\usepackage{enumitem}

\usepackage[pagebackref=true,breaklinks=true,letterpaper=true,colorlinks,bookmarks=false]{hyperref}

\usepackage[capitalize]{cleveref}
\crefname{section}{Sec.}{Secs.}
\Crefname{section}{Section}{Sections}
\Crefname{table}{Table}{Tables}
\crefname{table}{Tab.}{Tabs.}

\iccvfinalcopy 

\def\iccvPaperID{5594} 

\ificcvfinal\fi

\newcommand*{\affaddr}[1]{#1} 
\newcommand*{\affmark}[1][*]{\textsuperscript{#1}}

\begin{document}

\title{Blind2Sound: Self-Supervised Image Denoising without Residual Noise}

\author{Zejin Wang\affmark[1,2]\qquad Jiazheng Liu\affmark[1,3]\qquad Hao Zhai\affmark[1,3]\qquad Hua Han\affmark[1,3,]\thanks{Corresponding author}\\
 \affaddr{\normalsize\affmark[1]Institute of Automation, Chinese Academy of Sciences}\\
 \affaddr{\normalsize\affmark[2]School of Artificial Intelligence, University of Chinese Academy of Sciences}\\
 \affaddr{\normalsize\affmark[3]School of Future Technology, University of Chinese Academy of Sciences}\\
}
\maketitle
\ificcvfinal\thispagestyle{empty}\fi

\begin{abstract}
  Self-supervised blind denoising for Poisson-Gaussian noise remains a challenging task. Pseudo-supervised pairs constructed from single noisy images re-corrupt the signal and degrade the performance. The visible blindspots solve the information loss in masked inputs. However, without explicitly noise sensing, mean square error as an objective function cannot adjust denoising intensities for dynamic noise levels, leading to noticeable residual noise. In this paper, we propose Blind2Sound, a simple yet effective approach to overcome residual noise in denoised images. The proposed adaptive re-visible loss senses noise levels and performs personalized denoising without noise residues while retaining the signal lossless. The theoretical analysis of intermediate medium gradients guarantees stable training, while the Cramer Gaussian loss acts as a regularization to facilitate the accurate perception of noise levels and improve the performance of the denoiser. Experiments on synthetic and real-world datasets show the superior performance of our method, especially for single-channel images. The code is available in supplementary materials.
\end{abstract}

\begin{figure}[t]
\centering
\setlength{\abovecaptionskip}{0.1cm} 
\includegraphics[width=.48\textwidth]{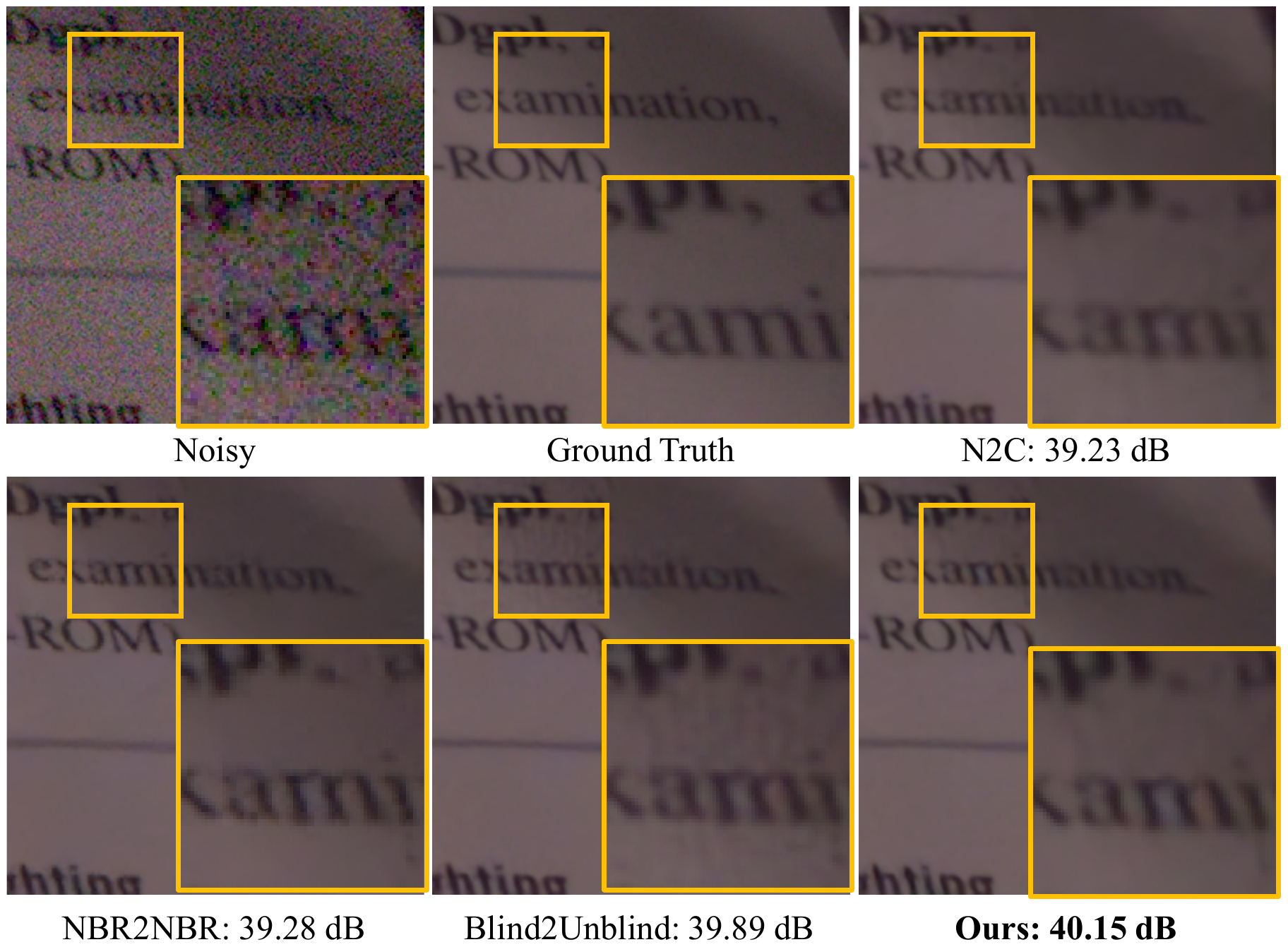}
   \caption{\textbf{Real-world image denoising on the SIDD validation.} Our approach performs thorough noise removal, while Blind2Unblind has considerable noise artifacts.}
\label{fig:val}
\vspace{-0.4cm}
\end{figure}

\section{Introduction}
\label{sec:intro}
Supervised image denoisers~\cite{ronneberger2015u,zhang2017beyond,zhang2018ffdnet,guo2019toward,anwar2019real,yue2019variational,zamir2020learning} have demonstrated impressive and superior performance by utilizing numerous noisy-clean pairs. Notably, several researchers~\cite{ronneberger2015u,zhang2017beyond,zhang2018ffdnet} pioneered the removal of additive white Gaussian noise and achieved notable performance gains. These denoisers~\cite{guo2019toward,anwar2019real,yue2019variational,zamir2020learning} can also eliminate other noise patterns, especially signal-dependent noise in the real world. However, acquiring a large amount of paired data to train a denoiser can be costly and, in some cases, such as CT and MRI, even impossible. Moreover, the performance of these denoisers rapidly degrades once they encounter unknown noise patterns due to the inherent prior embedded in the training data.

To tackle the challenge of acquiring paired data, researchers have conducted numerous studies~\cite{brooks2019unprocessing,zamir2020cycleisp,chen2018image,wu2020unpaired,lehtinen2018noise2noise,krull2019noise2void,batson2019noise2self} in two categories: data synthesis and self-supervised denoisers. Data synthesis involves generating paired data for supervised training via adding noise to clean images. For instance, UPI~\cite{brooks2019unprocessing} and CycleISP~\cite{zamir2020cycleisp} analyze the operations of the camera imaging pipeline on signal-dependent noise and create an imaging framework capable of generating arbitrary real image pairs in both raw-RGB and sRGB space. Other researchers~\cite{chen2018image,wu2020unpaired} synthesize noisy-clean image pairs for training using pre-estimated noise parameters. However, while data synthesis approaches are effective with limited specific data, their poor generality limits their broader application.

Noise2Noise~\cite{lehtinen2018noise2noise}, a special case of supervised denoising that uses corrupted image pairs, serves as the foundation for self-supervised denoising. Self-supervised denoisers, which learn only from a single raw noise image, have become the leading solution for denoising without data collection limitations. Blindspot schemes, including manual masking~\cite{krull2019noise2void,batson2019noise2self} and blindspot networks (BSN)~\cite{laine2019high,wu2020unpaired,cha2019fully,byun2020learning}, enable single image denoising by assuming that the signal is context-dependent and the noise is irrelevant. However, the information loss and sub-optimal masking strategies degrade the upper bound of denoising performance. Some works~\cite{xu2020noisy,moran2020noisier2noise,pang2021recorrupted} introduce additional noise to construct training data pairs. The pre-calibrated noise level restricts their application, and the re-corruption worsens the loss of valuable signals. NBR2NBR~\cite{huang2021neighbor2neighbor} sub-samples low-resolution training pairs from single noisy images for supervised training. The surrounding pixel approximation results in over-smoothing and block artifacts. FBI-Denoiser (FBI-D)~\cite{byun2021fbi} reimplements Gaussian estimation~\cite{chen2015efficient} as tensor operations combined with Generalized Anscombe Transformations (GAT)~\cite{anscombe1948transformation} to formulate Gaussian loss. However, Gaussian loss without fine-grained constraints zooms noise estimation errors. AP-BSN~\cite{lee2022ap} breaks spatial correlation but remains in the scope of BSN. Blind2Unblind~\cite{wang2022blind2unblind} achieves lossless denoising using the re-visible transition, but the greedy pixel-level fitting without noise sensing leads to sizable residual noise.

In this paper, we present a novel framework named Blind2Sound for self-supervised blind denoising, which achieves personalized denoising without noise residues while ensuring signal lossless. The framework includes an adaptive re-visible loss for lossless and personalized denoising and a Cramer Gaussian loss as a regularization to accurately sense noise knowledge. The framework assumes that the masked and visible branches are two independent generative processes, while denoising results following normal distributions and the respective noise following a Poisson-Gaussian distribution. This independence ensures that the masked results do not suppress the visible denoising. For optimal denoising, the adaptive loss adjusts noise removal via noise levels, while compatible with the lossless re-visible framework. The adaptive loss consists of two branches and their noise parameters, and the masked branch as an intermediate medium for gradient update. Due to signal dependence, the bottleneck of the masked branch greatly suppresses the accuracy of Poisson noise. Thus, the weighted Poisson noises in the mixed marginal likelihood is simplified to a single Poisson component. Moreover, the Cramer Gaussian loss introduces fine-grained noise knowledge, including sub-block and cross-channel constraints, to boost the estimation accuracy.

Our contribution can be summarized in three aspects:
\begin{enumerate}[itemsep=2pt,topsep=0pt,parsep=0pt]
\item The proposed self-supervised denoising framework adjusts denoising intensities based on sensed noise levels, achieving personalized denoising without residual noise while the signal is lossless.
\item We provide theoretical analysis for adaptive re-visible loss and the gradient of intermediate medium.
\item Our approach shows excellent performance on various real-world datasets with dynamic noise patterns.
\end{enumerate}

\section{Related Work}
\noindent\textbf{Self-Supervised Blind Image Denoising\;} 
Although Noise2Noise~\cite{lehtinen2018noise2noise} makes collecting paired data much easier, it remains impractical as it requires noisy image pairs from the same source. Therefore, blindspot schemes such as manual masking~\cite{krull2019noise2void,batson2019noise2self} and blindspot networks~\cite{laine2019high,cha2019fully} have been developed to achieve self-supervised denoising. However, these schemes suffer from the loss of valuable information, leading to severe artifacts. To address this issue, several methods, such as Laine19~\cite{laine2019high}, DBSN~\cite{wu2020unpaired}, FC-AIDE~\cite{cha2019fully}, BP-AIDE~\cite{byun2020learning}, and AP-BSN~\cite{lee2022ap}, have been proposed to enhance BSN. However, these methods either have poor performance or slow inference on Poisson-Gaussian noise. FBI-D~\cite{byun2021fbi} proposes a fast and effective blind image denoiser for Poisson-Gaussian noise, but its noise estimator has low accuracy in sRGB space without fine-grained noise constrains. Besides, the post-processing step can reinforce denoising errors accumulated in previous steps. Blind2Unblind~\cite{wang2022blind2unblind} overcomes these limitations and achieves fast lossless denoising under blindspots. However, implicit denoising without sensing noise levels is unable to provide personalized noise removal for Poisson-Gaussian noise, resulting in residual noise artifacts.

\noindent\textbf{Noise Estimation Methods\;} 
Current noise estimation methods~\cite{liu2013single,pyatykh2012image,chen2015efficient,foi2008practical,liu2014practical} are mainly designed for two noise models: additive white Gaussian noise (AWGN) and Poisson-Gaussian noise. For AWGN, a signal-independent noise, once the noise variance is known, its distribution can be determined. Previous methods~\cite{liu2013single,pyatykh2012image} for estimating AWGN variance used low-rank patch selection and principal component analysis. Recently, Chen~\etal~\cite{chen2015efficient} addressed the bias issue for Gaussian parameter estimation using statistical decomposition of eigenvalues. In contrast, Poisson-Gaussian noise~\cite{foi2008practical} is a mixture of Poisson and Gaussian components and is often considered as source-related noise in real-world scenarios. Foi~\etal~\cite{foi2008practical} introduced the Poisson-Gaussian model and provided fully automated noise parameter estimation for a single noisy image. More recently, Liu~\etal~\cite{liu2014practical} improved local mean and noise variance estimation for selected low-rank patches.

\begin{figure*}[ht]
\centering
\vspace{-0.15cm}
\includegraphics[width=.91\textwidth]{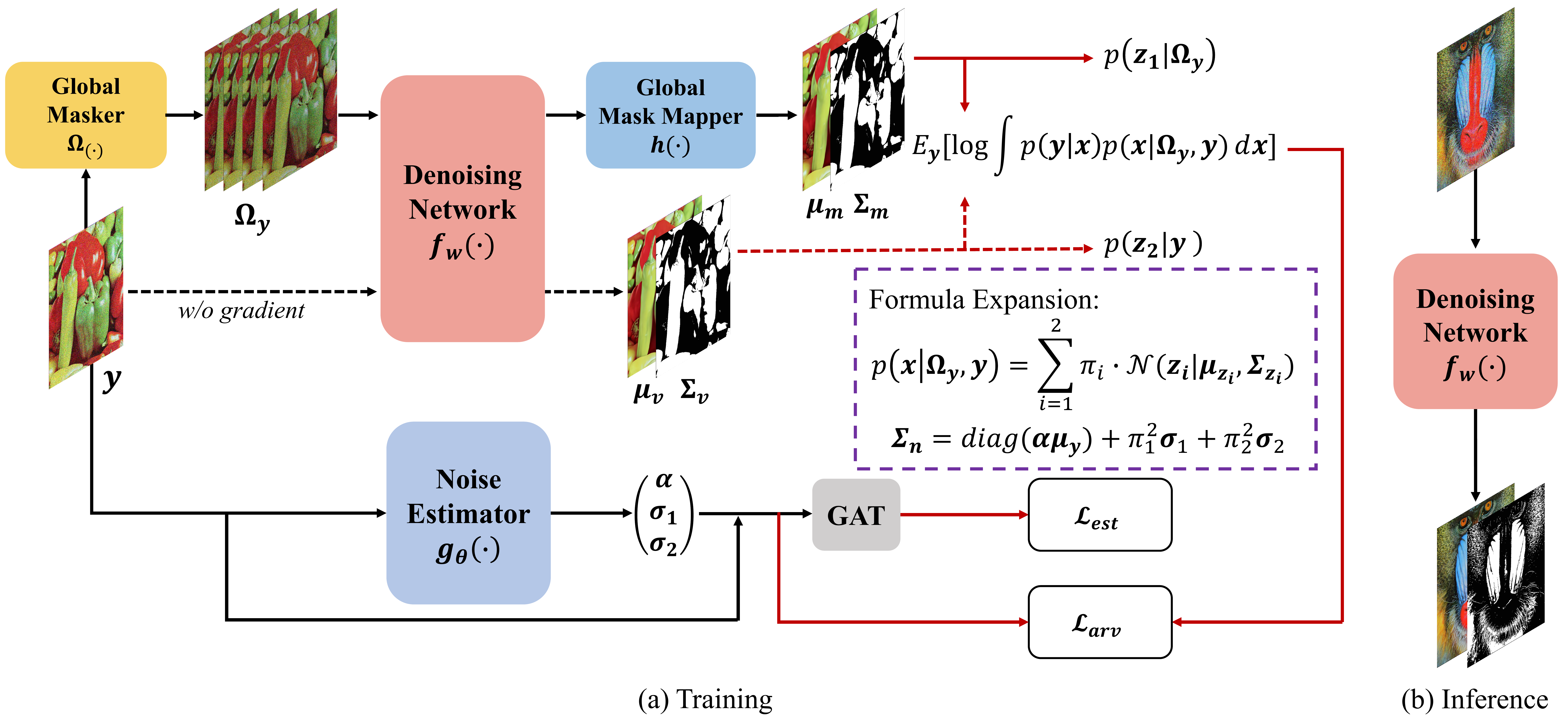}
   \caption{The architecture of the proposed Blind2Sound framework. The masked volume is represented by $\boldsymbol{\Omega_{y}}$, and the dotted lines indicate the forward process without updating the gradient. The Global Masker and Global Mask Mapper are detailed in the Supplementary Material (S.M.). Note that the noise estimator and the denoiser are trained through joint optimization. During inference, the auxiliary branches, namely the noise estimator and masked branch, are removed, and the denoiser directly generates denoised images from raw noisy inputs $\boldsymbol{y}$.}
\label{fig:overview}
\vspace{-0.3cm}
\end{figure*}

\section{Problem Setting and Preliminaries}
\subsection{Problem Formulation}
Given a noisy observation $\boldsymbol{y}$, our goal is to learn the clean image $\boldsymbol{x}$ and its noise parameters directly from a single noisy image. We focus on Poisson-Gaussian noise~\cite{foi2008practical}, which is common in real-world imaging sensors. The noise-corrupted observation $\boldsymbol{y}$ can be described as follows:

\begin{equation}
\label{eq:1}
\boldsymbol{y} = \alpha P + N,
\end{equation}
where $P\sim \rm{Poisson}(\boldsymbol{x}/\alpha)$ is the signal-dependent Poisson noise caused by photon counting, and $N\sim \mathcal{N}(0,\sigma^2)$ is the signal-independent Gaussian noise resulting from electric and thermal noise. Here, $\alpha$ is a scaling factor that depends on the quantum efficiency and analog gain. To simplify the problem, the Poisson noise is approximated as signal-independent Gaussian noise~\cite{hasinoff2014photon}, and the corruption can be reformulated as:
\begin{equation}
\label{eq:2}
\boldsymbol{y} = \boldsymbol{x} + \mathcal{N}(0,\alpha\boldsymbol{x}+\sigma^2).
\end{equation}

\subsection{Generalized Anscombe Transformation (GAT)}
GAT~\cite{anscombe1948transformation} is a popular variance-stabilized transform used for Poisson-Gaussian noise. It converts the mixture noise into stable Gaussian noise with unit variance:
\begin{equation}
\label{eq:3}
G_{\alpha,\sigma}(\boldsymbol{y}) = \frac{2}{\alpha}\sqrt{\alpha\boldsymbol{y}+\frac{3}{8}\alpha^2+\sigma^2}.
\end{equation}

\subsection{Gaussian Loss}
As discussed in~\Cref{sec:intro}, Byun~\etal~\cite{byun2021fbi} introduce a noise estimation network, denoted as $h_{\boldsymbol{\theta}}(\cdot)$, that takes Poisson-Gaussian noise-corrupted images $\boldsymbol{y}$ as input and produces the estimated noise parameters $(\hat{\alpha},\hat{\sigma})$. To train this network, they reimplement the Gaussian parameter estimation proposed in~\cite{chen2015efficient} as tensor operations $\eta(\cdot)$ and leverage the property that the transformed noise level using GAT should be closed to unit variance. Therefore, the Gaussian loss is defined for noise estimation as:
\begin{equation}
\label{eq:4a}
\mathop{\arg\min}\limits_{\boldsymbol{\theta}} \mathbb{E}_{\mathbf{y}}\left\Vert \eta\left(G_{\hat{\alpha}\left(\boldsymbol{\theta}\right),\hat{\sigma}\left(\boldsymbol{\theta}\right)}\left(\boldsymbol{y}\right)\right) - 1\right\Vert^2_2,
\end{equation}
the noise estimation network $h_{\boldsymbol{\theta}}(\boldsymbol{y})$ predicts the noise parameters $\hat{\alpha}\left(\boldsymbol{\theta}\right)$ and $\hat{\sigma}\left(\boldsymbol{\theta}\right)$. Then, the transformed noisy image with unit variance is denoised and the output of BSN is mapped back to the original image using the Inverse Anscombe transformation (IAT)~\cite{makitalo2012optimal}. Note that noise estimation and denoising are trained separately without joint optimization.

\subsection{Blind2Unblind Revisit}
Blind2Unblind~\cite{wang2022blind2unblind} is a self-supervised noise removal method that achieves lossless denoising using the re-visible transition. The overall framework consists of two branches: a masked branch that uses pseudo-supervised pairs for auxiliary denoising and a visible branch that learns directly from raw noise images. Given the noisy observation $\mathbf{y}$, Blind2Unblind minimizes the following re-visible loss:
\begin{equation}
\label{eq:4b}
\mathop{\arg\min}\limits_{\theta} \mathbb{E}_{\mathbf{y}}\Vert h(f_{\theta}(\mathbf{\Omega_{\mathbf{y}}}))+\lambda\hat{f}_{\theta}(\mathbf{y})-(\lambda+1)\mathbf{y}\Vert^2_2,
\end{equation}
where $f_{\theta}(\cdot)$ denotes the neural network with parameter $\theta$, $\hat{f}_{\theta}(\cdot)$ indicates that the gradient is disabled. $\mathbf{\Omega_{y}}$ is a noisy masked volume where blind spots cover the noisy image $\mathbf{y}$. The function $h(\cdot)$ samples denoised pixels at the blind spots, and $\lambda$ is a visible constant. The re-visible transition refers to the smooth transition of the objective function from blind denoising to visible denoising as $\lambda$ increases.

\section{Main Method}
\label{sec:bbd}
As illustrated in~\Cref{fig:overview}, the training process involves two modules: 1) The denoising network $f_{\omega}(\cdot)$ that outputs denoised results of the noisy masked volume $\boldsymbol{\Omega_{y}}$ and the raw noisy image $\boldsymbol{y}$, including their mean $\boldsymbol{\mu_{m}}, \boldsymbol{\mu_{v}}$, and covariance $\boldsymbol{\Sigma_{m}}, \boldsymbol{\Sigma_{v}}$. 2) The noise estimator $g_{\theta}(\cdot)$ that calculates the Poisson-Gaussian noise parameters $(\alpha, \sigma_1, \sigma_2)$. Since actual noise levels of denoised images may differ from raw noisy inputs, Cramer Gaussian loss only serves as a regularization to assist the denoiser in picking appropriate noise levels. Namely, the adaptive re-visible loss also determines the noise parameters for explicit and lossless personalized denoising. The training set $D$ contains $n$ training images denoted as $D=\{\boldsymbol{y}_{i}\}_{i=1}^{n}$, where $\boldsymbol{y}_{i}$ denotes the $i^{th}$ raw noisy image. The details of the proposed method are presented in subsequent sections.

\subsection{Motivation}
The re-visible framework overcomes the information loss in blindspot-driven methods. However, mean square error as objective function cannot adapt to varying noise levels, frequently resulting in residual noise in denoised images. Therefore, it is necessary to design an adaptive loss that adjusts denoising intensities based on sensed noise levels. For optimal noise removal, the adaptive loss should personalized denoising without noise residues, while ensuring valuable signal lossless. Lossless denoising requires that the adaptive loss is compatible with the re-visible framework. For this purpose, we regard the masked and visible branches as Gaussian processes and design an adaptive re-visible loss that satisfies both personalized and lossless requirements.

\subsection{Adaptive Re-Visible Loss}
The goal of adaptive re-visible loss is to achieve optimal noise removal while retaining lossless denoising. We achieve this by re-considering the re-visible scheme from Bayesian reasoning and developing a personalized noise removal method that does not require an auxiliary branch for noise estimation during inference. 

First, we model $p(\boldsymbol{z_1}|\boldsymbol{\Omega_y})$ as a multivariate Gaussian which represents that the latent clean image $\boldsymbol{z_1}$ is generated from the masked noisy volume $\boldsymbol{\Omega_{y}}$ as follows: 
\begin{equation}
\label{eq:5}
\boldsymbol{z_1} \sim \mathcal{N}(\boldsymbol{z_1}|\boldsymbol{\mu_m},\boldsymbol{\Sigma_m}),
\end{equation}
where $\mathcal{N}(\cdot|\boldsymbol{\mu_m},\boldsymbol{\Sigma_m})$ denotes the multivariate Gaussian distribution with mean $\boldsymbol{\mu_m}$ and variance $\boldsymbol{\Sigma_m}$.

For the masked branch,~\cref{eq:2} incorporates extra noise knowledge into the explicit corruption model, provided as the likelihood $p(\boldsymbol{y}|\boldsymbol{z_1})$ given a clean value. Therefore, the marginal likelihood of the noisy training data can be constructed via the distribution of unobserved clean data $\boldsymbol{z_1}$:
\begin{equation}
\label{eq:6}
p(\boldsymbol{y_1})=\int p(\boldsymbol{y}|\boldsymbol{z_1})p(\boldsymbol{z_1}|\boldsymbol{\Omega_y})d\boldsymbol{z_1}.
\end{equation}

As illustrated in~\cref{eq:6}, when only noisy training data $\boldsymbol{y}$ are available, a known noise model is able to explicitly predict the masked prior $p(\boldsymbol{z_1}|\boldsymbol{\Omega_y})$. Specifically, for an approximate Gaussian noise model, the covariance of two mutually independent Gaussian convolutions is simply the sum of the components~\cite{bromiley2003products}. Hence, the marginal likelihood $p(\boldsymbol{y_1})$ is calculated in closed form, allowing to obtain the distribution of $\boldsymbol{z_1}$ by maximizing~\cref{eq:6}. According to~\cref{eq:2,eq:5} as well as the above analysis, the mean and variance of $\boldsymbol{y_1}$ become: 
\begin{equation}
\label{eq:7}
\boldsymbol{y_1} \sim \mathcal{N}(\boldsymbol{y_1}|\boldsymbol{\mu_m},\boldsymbol{\Sigma_m}+diag(\alpha_1\boldsymbol{\mu_m})+\sigma_1\boldsymbol{\rm{I}}).
\end{equation}
For the visible branch, we construct $p(\boldsymbol{z_2}|\boldsymbol{y})$ as the generation of the latent clean image $\boldsymbol{z_2}$ from the raw noise image $\boldsymbol{y}$, which then becomes:
\begin{equation}
\label{eq:8}
\boldsymbol{z_2} \sim \mathcal{N}(\boldsymbol{z_2}|\boldsymbol{\mu_v},\boldsymbol{\Sigma_v}),
\end{equation}
where the mean $\boldsymbol{\mu_v}$ and variance $\boldsymbol{\Sigma_v}$ are directly generated from the raw noise image $\boldsymbol{y}$ without gradient update. The marginal likelihood for the visible branch via the distribution of unobserved clean data $\boldsymbol{z_2}$ is then formulated as:
\begin{equation}
\label{eq:9}
p(\boldsymbol{y_2})=\int p(\boldsymbol{y}|\boldsymbol{z_2})p(\boldsymbol{z_2}|\boldsymbol{y})d\boldsymbol{z_2}.
\end{equation}
Similar to~\cref{eq:7}, the mean and variance of $\boldsymbol{y_2}$ become:
\begin{equation}
\label{eq:10}
\boldsymbol{y_2} \sim \mathcal{N}(\boldsymbol{y_2}|\boldsymbol{\mu_v},\boldsymbol{\Sigma_v}+diag(\alpha_2\boldsymbol{\mu_v})+\sigma_2\boldsymbol{\rm{I}}).
\end{equation}

The marginal likelihood $p(\boldsymbol{y_1})$ for the blind branch and $p(\boldsymbol{y_2})$ for the visible branch are now available. The mean and variance of $\boldsymbol{z_2}$ do not participate in backpropagation due to identity mapping. However, maximizing only the blind distribution $p(\boldsymbol{z_1}|\boldsymbol{\Omega_y})$ via~\cref{eq:6} has limited performance. To improve the performance, the loss errors of the visible branch $p(\boldsymbol{z_2}|\boldsymbol{y})$ are incorporated into the mask gradient. The decorrelation of the two branches enhances visible denoising without suppressing the masked results. Thus, $\boldsymbol{y_1}$ and $\boldsymbol{y_2}$ are modeled as i.i.d., and a Gaussian mixture is applied to boost their representation. Combining~\cref{eq:7,eq:10}, an enhanced target distribution $\boldsymbol{y}$ is obtained while retaining the independence of the two branches:
\begin{equation}
\label{eq:11}
\boldsymbol{y} \sim \sum\limits_{i=1}^{2}\pi_{i}\cdot\mathcal{N}(\boldsymbol{y_i}|\boldsymbol{\mu_{y_i}},\boldsymbol{\Sigma_{y_i}}),
\end{equation}
where $\pi_i$ is a hyper-parameter for the degree of re-visible. Besides, $0\leq\pi_i\leq1$ and $\pi_1+\pi_2=1$. Set the blind factor $\pi_1$ to ${1}/{(1+\lambda)}$, the visible factor $\pi_2$ to ${\lambda}/{(1+\lambda)}$ and $\lambda$ is a growing constant. Then,~\cref{eq:11} is reformulated as:
\begin{equation}
\label{eq:12}
\boldsymbol{y} \sim \mathcal{N}(\boldsymbol{y}|\frac{\boldsymbol{\mu_m}+\lambda\boldsymbol{\mu_v}}{1+\lambda},\frac{\boldsymbol{\Sigma_{y_1}}+\lambda^2\boldsymbol{\Sigma_{y_2}}}{(1+\lambda)^2}).
\end{equation}
To simplify notation, we use $\boldsymbol{y} \sim \mathcal{N}(\boldsymbol{y}|\boldsymbol{\mu_y},\boldsymbol{\Sigma_y})$ and denote the clean target image as $\boldsymbol{x}$. The mask mean $\boldsymbol{\mu_m}$ is a lower bound of $\boldsymbol{x}$, and the signal-dependent factor $\alpha_1$ magnifies this error. To improve accuracy, we replace the noise model in~\cref{eq:12} with a more precise $p(\boldsymbol{y}|\boldsymbol{x})$ that has zero mean and variance $diag(\alpha\boldsymbol{\mu_y})+\pi^2_1\sigma_1\boldsymbol{\rm{I}}+\pi^2_2\sigma_2\boldsymbol{\rm{I}}$. The enhanced mixture marginal likelihood that bridges the blind and visible branches then becomes:
\begin{equation}
\label{eq:13}
p(\boldsymbol{y})=\int p(\boldsymbol{y}|\boldsymbol{x})p(\boldsymbol{x}|\boldsymbol{y},\boldsymbol{\Omega_y})d\boldsymbol{x}.
\end{equation}
To fit the observed noisy training data, we minimize its negative log-likelihood loss in the training phase as follows:
\begin{equation}
\label{eq:14}
\begin{split}
\mathcal{L}_{arv}&=-\log p(\boldsymbol{y})=-\log\left[\pi_1p(\boldsymbol{y_1})+\pi_2p(\boldsymbol{y_2})\right]\\
                 &=\frac{1}{2}[(\boldsymbol{y}-\boldsymbol{\mu_y})^{T}\boldsymbol{\Sigma_y}^{-1}(\boldsymbol{y}-\boldsymbol{\mu_y})]\\
                 &\quad+\frac{1}{2}\log|\boldsymbol{\Sigma_y}|+const,
\end{split}
\end{equation}
where $const$ is an additive constant term that can be discarded, $\mathcal{L}_{arv}$ denotes the proposed adaptive re-visible loss. When the denoiser converges, the following is the optimal clean value $\boldsymbol{\tilde{x}}$ of~\cref{eq:14}:
\begin{equation}
\label{eq:15}
\boldsymbol{\tilde{x}}=\frac{\boldsymbol{\mu_m}+\lambda\boldsymbol{\mu_v}}{1+\lambda}.
\end{equation}
Assuming that $\boldsymbol{\mu_m}=\boldsymbol{x}+\varepsilon_1$ and $\boldsymbol{\mu_v}=\boldsymbol{x}+\varepsilon_2$. Empirically, $\Vert\varepsilon_1\Vert_1 > \Vert\varepsilon_2\Vert_1$ because valuable information is lost in $\boldsymbol{\Omega_y}$. Combined with~\cref{eq:11,eq:15}, it can be concluded that $\boldsymbol{\mu_m}\leq \boldsymbol{\tilde{x}}\leq \boldsymbol{\mu_v}$ since $\boldsymbol{\tilde{x}}$ is the weighted average of $\boldsymbol{\mu_m}$ and $\boldsymbol{\mu_v}$. 

Using the analytic form of adaptive re-visible loss in~\cref{eq:14}, we explore and confirm the collaborative mechanism between the gradient update medium $\boldsymbol{\mu_m}$ and the visible constant $\boldsymbol{\mu_v}$. Let $\boldsymbol{n_y}=\boldsymbol{y}-\boldsymbol{\mu_y}$, the derivative of the medium $\boldsymbol{\mu_m}$ gives its gradient $\nabla_{\boldsymbol{\mu_m}}$ (see S.M.):
\begin{small}
\begin{equation}
\label{eq:16}
\begin{aligned}
\frac{\partial\mathcal{L}_{arv}}{\partial\boldsymbol{\mu_m}}&=-\frac{1}{1+\lambda}\boldsymbol{\Sigma_y}^{-1}\boldsymbol{n_y}+\frac{\alpha}{(\lambda+1)}(\boldsymbol{\Sigma_y}^{-1}-\\&\quad\;\boldsymbol{\Sigma_y}^{-1}\boldsymbol{n_y}\boldsymbol{n_y}^{T}\boldsymbol{\Sigma_y}^{-1})\odot\boldsymbol{\rm{I}}\odot(\sqrt{\boldsymbol{\mu_y}\boldsymbol{\mu_y}^{T}})'\boldsymbol{\mu_y}.
\end{aligned}
\end{equation}
\end{small}

According to~\cref{eq:16}, it is observed that including the gradient update term $\boldsymbol{\mu_m}$ in $\rm{diag}(\alpha\boldsymbol{\mu_y})$ results in severe instability during training due to a complicated second term in the gradient. Therefore, disabling the gradient of $\rm{diag}(\alpha\boldsymbol{\mu_y})$ is considered to stabilize the training process and improves the performance of the denoiser. Moreover, denoising can be performed directly from the raw noise image $\boldsymbol{y}$ during inference.

Laine19~\etal~\cite{laine2019high} utilize Bayesian reasoning to incorporate information from $\boldsymbol{y}$ into maximum posterior probabilities (MAP) during test time. However, post-processing, which is not involved in training, performs poorly in practice. In contrast, additional MAP is redundant for adaptive re-visible denoising as $p(\boldsymbol{x}|\boldsymbol{\Omega_y},\boldsymbol{y})$ in~\cref{eq:13} already includes information from $\boldsymbol{y}$. As a result, our approach outperforms other self-supervised methods.

\subsection{Cramer Gaussian Loss}
Gaussian loss~\cite{byun2021fbi} estimates noise directly from global images, which ignores the perceptual dimension of local noise knowledge. As a result, the estimated Poisson Gaussian parameters are less accurate. To overcome the coarse-scale limitation, the proposed Cramer Gaussian loss uses fine-grained local sub-block or cross-channel constraints to enrich noise perception dimensions, reducing the solution space to find the exact median of the noise level.  

For single-channel images without channel correlation, each sub-block of the GAT-transformed images should have same noise levels as global ones. To ensure this, we introduce a fine-grained sub-block noise level constraint based on the coarse-scale constraint. The estimation incorporates overlapping sub-blocks at four corners as local noise knowledge for single-channel images. Denote $g_{\theta}(\boldsymbol{y})$ as the estimated noise parameter $(\hat{\alpha}, \hat{\sigma})$, $\eta(\cdot)$ as the Gaussian estimator~\cite{chen2015efficient} that estimates the Gaussian noise variance, and $G_{g_{\theta}}(\boldsymbol{y})$ as the GAT-transformed image $G_{g_{\theta}(\boldsymbol{y})}(\boldsymbol{y})$. The estimated Gaussian variance for the transformed noise and its sub-blocks should approximate unit variance. Cramer Gaussian loss for single-channel image $\boldsymbol{y}$ becomes:
\begin{small}
\begin{equation}
\label{eq:17}
\mathcal{L}_{est}= \sum\limits_{s=1}^{4} \left\Vert \eta\left(G_{g_{\theta}}(\boldsymbol{y}^{s})\right) - 1 \right\Vert^2_2 + \left\Vert \eta\left(G_{g_{\theta}}(\boldsymbol{y})\right) - 1 \right\Vert^2_2,
\end{equation}
\end{small}
where $\boldsymbol{y}^{s}_{i}$ denotes the $s^{th}$ sub-block cropped from $\boldsymbol{y}_{i}$. We crop four identical sub-blocks from four corners. Each sub-block is three-quarters the size of the original image.

For multi-channel images, the implied noise level should be the same for different channels. To ensure a unique solution space, we introduce cross-channel noise level approximation, enabling cross-channel information exchange and overcoming the inherent limitation of Gaussian loss,~\ie, the problem of inter-channel estimated error offsetting. For each channel in the GAT-transformed image, the Gaussian noise variance of each channel should approximate the unit variance, and the noise level should be the same for each channel. Thus, the Cramer Gaussian loss for the multi-channel image $\boldsymbol{y}$ becomes:
\begin{small}
\begin{equation}
\label{eq:18}
\begin{split}
\mathcal{L}_{est} &= \sum\limits_{j\ne k}^{c} \left\Vert \eta\left(G_{g_{\theta}}(\boldsymbol{y}_{j})\right) - 1 \right\Vert^2_2 \\
                  &+ \left\Vert \eta\left(G_{g_{\theta}}(\boldsymbol{y}_{j})\right) - \eta\left(G_{g_{\theta}}(\boldsymbol{y}_{k})\right) \right\Vert^2_2,
\end{split}
\end{equation}
\end{small}
where $c$ is the number of channels in image $\boldsymbol{y}_{i}$. $j,k$ represent the $j^{th}$ and $k^{th}$ channel, respectively.

\section{Experimental Results}
\subsection{Implementation Details}
\noindent\textbf{Training Details\;}
We use the same noise estimator as the FBI-D~\cite{byun2021fbi} and a modified U-Net~\cite{laine2019high,huang2021neighbor2neighbor,wang2022blind2unblind} as the denoising network. Adam~\cite{kingma2015adam} with a weight decay of $1e^{-8}$ is used as the optimizer. The initial learning rate for the noise estimator is $0.0001$. For a small training set, the initial learning rate of the denoising network is $0.001$, while $0.0003$ and $0.0001$ for ILSVRC2012~\cite{deng2009imagenet} validation set and SIDD~\cite{abdelhamed2018high}, respectively. The learning rate for the noise estimator decreases by half every $10$ epochs with $50$ epochs trained, while the learning rate for the denoising network is halved every $20$ epochs with $100$ epochs trained. As for the hyper-parameter in the adaptive re-visible loss, we set $\lambda=3$ as the initial value and progressively increase it to $11$. The denoising network and noise estimation are jointly optimized during training. The patches of size $128\times128$ are randomly cropped for training.
Note that the details of the global masker and global mask mapper remain the same as Blind2Unblind~\cite{wang2022blind2unblind}. All models are trained using Python 3.10.4, Pytorch 1.11.0~\cite{paszke2019pytorch}, and an Nvidia Tesla V100 GPU.

\noindent\textbf{Datasets\;}
We consider five types of Poisson-Gaussian noise in synthetic noise estimation: (1) $PG1:\alpha=0.1,\sigma=0.02$, (2) $PG2:\alpha=0.1,\sigma=0.0002$, (3) $PG3:\alpha=0.05,\sigma=0.02$, (4) $PG4:\alpha=0.05,\sigma=0.0002$, (5) $PG5:\alpha=0.01,\sigma=0.02$. For grayscale images, we use BSD400~\cite{martin2001database} for the training set, and for sRGB images, we use CBSD432~\cite{zhang2017beyond}. The noise levels are estimated on standard BSD68~\cite{roth2005fields} and CBSD68~\cite{roth2005fields} for grayscale and sRGB images, respectively. For synthetic denoising, we use ILSVRC2012~\cite{deng2009imagenet} validation set for sRGB image denoising and BSD400~\cite{martin2001database} for grayscale image denoising. Specifically, following the setting in~\cite{laine2019high,huang2021neighbor2neighbor,wang2022blind2unblind}, we select $44328$ images with sizes between $256\times256$ and $512\times512$ pixels from ILSVRC2012 validation set for training. The test sets used for grayscale image denoising are Set12, BSD68~\cite{roth2005fields}, and Urban100~\cite{huang2015single}, while for sRGB denoising, we use Kodak~\cite{franzen1999kodak}, BSD300~\cite{martin2001database}, and Set14~\cite{zeyde2010single}. For real noise experiments, we use SIDD~\cite{abdelhamed2018high} for real-world denoising in raw-RGB space and Fluorescence Microscopy Denoising (FMD)~\cite{zhang2018poisson} for real-world grayscale denoising. We train SIDD using only the raw-RGB images in the SIDD Medium Dataset and validate and test using the SIDD Validation and Benchmark Datasets. Note that we receive the evaluation results for the SIDD Benchmark from the online public website~\cite{web:siddbenchmark}. For FMD, we use the Confocal Mice and Two-Photon Mice datasets, and the 19th view is used for testing.

\noindent\textbf{Baselines\;}
We compare the proposed Cramer Gaussian loss with three noise estimation methods, including the Gaussian loss in FBI-D~\cite{byun2021fbi}, Foi~\cite{foi2008practical}, and Liu~\cite{liu2014practical}, and evaluate the denoising performance of Blind2Sound against two supervised methods (N2C~\cite{ronneberger2015u} and N2N~\cite{lehtinen2018noise2noise}), a traditional approach (GAT+BM3D~\cite{dabov2007image}), and four self-supervised algorithms (N2V~\cite{krull2019noise2void}, NBR2NBR~\cite{huang2021neighbor2neighbor}, FBI-D~\cite{byun2021fbi}, and Blind2Unblind~\cite{wang2022blind2unblind}). We adopt the experimental setting of FBI-D and Blind2Unblind and reproduce the results using the official implementation for a fair comparison.

\begin{table}[ht]
\scriptsize
  \centering
  \setlength{\abovecaptionskip}{0.1cm} 
  \setlength\tabcolsep{1.5pt} 
  \begin{tabular}[b]{@{}ccccccc@{}}
    \toprule
      \multirow{3}{*}[-0.5ex]{\shortstack[c]{Noise\\Level}}
      & \multicolumn{4}{c}{Grayscale} & \multicolumn{2}{c}{sRGB} \\
      \cmidrule(lr){2-5} \cmidrule(lr){6-7}
      & \multicolumn{1}{c}{Foi~\cite{foi2008practical}} & \multicolumn{1}{c}{Liu~\cite{liu2014practical}} & \multicolumn{1}{c}{FBI-D~\cite{byun2021fbi}} & \multicolumn{1}{c}{Ours} & \multicolumn{1}{c}{FBI-D~\cite{byun2021fbi}} & \multicolumn{1}{c}{Ours} \\
      & $(\hat{\alpha},\hat{\sigma})$ & $(\hat{\alpha},\hat{\sigma})$ & $(\hat{\alpha},\hat{\sigma})$ & $(\hat{\alpha},\hat{\sigma})$ & $(\hat{\alpha},\hat{\sigma})$ & $(\hat{\alpha},\hat{\sigma})$ \\

    \cmidrule(lr){1-1} \cmidrule(lr){2-5} \cmidrule(lr){6-7}
      $PG1$ & \textbf{0.096}/0.042 & 0.072/0.045 & 0.092/0.039 & 0.093/\textbf{0.019} & 0.080/\textbf{0.021} & \textbf{0.099}/0.001 \\
    
      $PG2$ & \textbf{0.097}/0.035 & 0.071/0.044 & 0.083/0.061 & 0.090/\textbf{0.001} & 0.074/0.033 & \textbf{0.096}/\textbf{0.001} \\
    
      $PG3$ & \textbf{0.049}/0.031 & 0.04/0.04 & 0.052/0.003 & 0.048/\textbf{0.022} & 0.041/0.015 & \textbf{0.051}/\textbf{0.020} \\
   
      $PG4$ & \textbf{0.051}/0.018 & 0.039/0.034 & 0.046/0.035 & 0.048/\textbf{0.013} & 0.040/0.013 & \textbf{0.050}/\textbf{0.0007} \\

      $PG5$ & 0.011/0.027 & 0.007/0.032 & 0.009/0.034 & \textbf{0.010}/\textbf{0.021} & 0.008/0.009 & \textbf{0.010}/\textbf{0.028} \\
    \bottomrule
  \end{tabular}
  \caption{The average noise parameter estimates for grayscale images in BSD68 and sRGB images in CBSD68.}
  \label{tab:est}
  \vspace{-0.4cm}
\end{table}

\subsection{Results for Noise Estimation}
We first evaluate the Cramer Gaussian loss for noise estimation.~\Cref{tab:est} shows the average noise parameters $(\hat{\alpha}, \hat{\sigma})$ predicted by Foi~\cite{foi2008practical}, Liu~\cite{liu2014practical}, FBI-D~\cite{byun2021fbi}, and Cramer Gaussian loss for five noise patterns in grayscale and sRGB images. Cramer Gaussian loss shows superior performance in estimating Gaussian parameters for grayscale images, and its Poisson level estimation is comparable to Foi while requiring less inference time. Moreover, compared to FBI-D, Cramer Gaussian loss eliminates its severe Gaussian estimation error on grayscale images by using a fine-grained fusion strategy. For sRGB space, cross-channel approximation enables our method to predict Poisson-Gaussian parameters close to their actual value, whereas FBI-D is inaccurate for estimating noise in sRGB images.

\begin{figure*}[t]
\centering
\vspace{-0.2cm}
\setlength{\abovecaptionskip}{0.cm} 
\includegraphics[width=\textwidth]{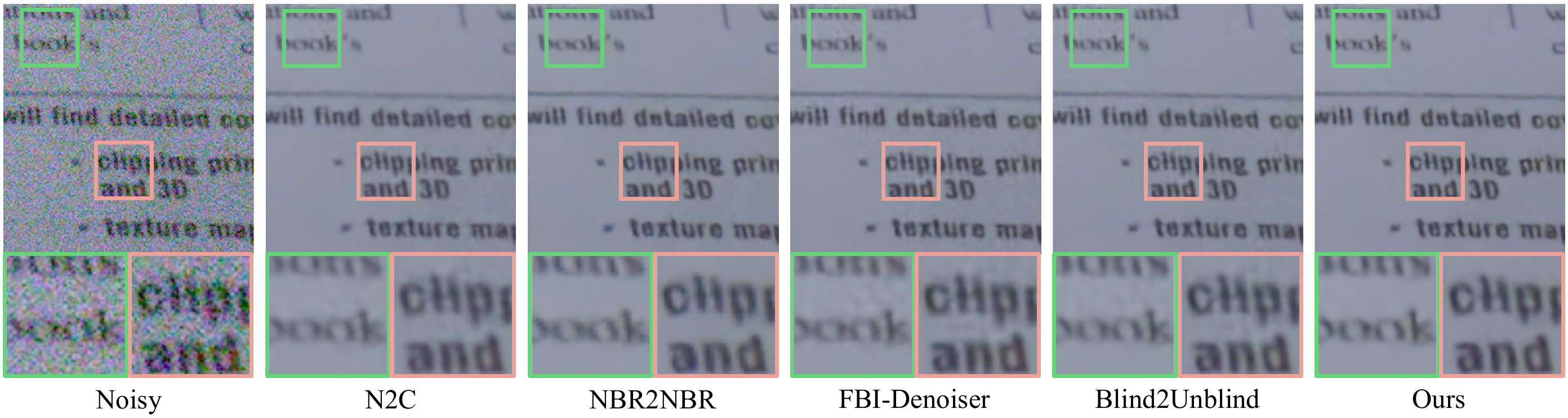}
   \caption{Visualization results on one typical image of the SIDD Benchmark.}
\label{fig:sidd}
\vspace{-0.5cm}
\end{figure*}

\begin{table}[ht]
\scriptsize
  \centering
  \setlength\tabcolsep{4pt} 
  \begin{tabular}[b]{clccc}
    \toprule
      Noise Type & Method & BSD68 & Set12 & Urabn100 \\
    \midrule
      \multirow{10}{*}{\shortstack[c]{$\alpha=0.01$\\$\sigma=0.0002$}} 
        & Baseline, N2C~\cite{ronneberger2015u} & 30.82/0.877 & 31.49/0.881 & 30.80/0.901 \\
        & Baseline, N2N~\cite{lehtinen2018noise2noise} & 30.76/0.876 & 31.47/0.880 & 30.84/0.900 \\
        \cline{2-5}
        & GAT+BM3D~\cite{dabov2007image} & 30.08/0.849 & 30.36/0.871 & 30.44/0.881 \\ 
        & N2V~\cite{krull2019noise2void} & 29.04/0.824 & 29.29/0.830 & 27.48/0.830 \\
        & FBI-D~\cite{byun2021fbi} & 28.09/0.784 & 29.13/0.828 & 27.28/0.829 \\
        & NBR2NBR~\cite{huang2021neighbor2neighbor} & 30.38/0.861 & 31.07/0.869 & 30.15/0.884 \\
        & Blind2Unblind~\cite{wang2022blind2unblind} & \underline{30.61}/\underline{0.869} & \underline{31.45}/\underline{0.880} & \underline{30.70}/\underline{0.900} \\
        & Ours & \textbf{30.83}/\textbf{0.875} & \textbf{31.68}/\textbf{0.886} & \textbf{31.14}/\textbf{0.908} \\      
    \midrule
      \multirow{10}{*}{\shortstack[c]{$\alpha=0.01$\\$\sigma=0.02$}} 
        & Baseline, N2C~\cite{ronneberger2015u} & 30.54/0.868 & 31.26/0.876 & 30.51/0.892 \\
        & Baseline, N2N~\cite{lehtinen2018noise2noise} & 30.62/0.871 & 31.34/0.878 & 30.70/0.899 \\
        \cline{2-5}
        & GAT+BM3D~\cite{dabov2007image} & 29.86/0.844 & 30.55/0.857 & 30.23/0.882 \\ 
        & N2V~\cite{krull2019noise2void} & 29.01/0.828 & 29.43/0.842 & 27.53/0.838 \\
        & FBI-D~\cite{byun2021fbi} & 27.95/0.776 & 29.11/0.824 & 27.21/0.822 \\
        & NBR2NBR~\cite{huang2021neighbor2neighbor} & 30.28/0.860 & 31.00/0.868 & 30.15/0.886 \\
        & Blind2Unblind~\cite{wang2022blind2unblind} & \underline{30.41}/\underline{0.865} & \underline{31.27}/\underline{0.877} & \underline{30.49}/\underline{0.896} \\
        & Ours & \textbf{30.62}/\textbf{0.869} & \textbf{31.47}/\textbf{0.880} & \textbf{30.91}/\textbf{0.902} \\    
    \midrule
      \multirow{10}{*}{\shortstack[c]{$\alpha=0.05$\\$\sigma=0.02$}} 
        & Baseline, N2C~\cite{ronneberger2015u} & 27.11/0.765 & 27.78/0.801 & 26.69/0.813 \\
        & Baseline, N2N~\cite{lehtinen2018noise2noise} & 27.07/0.762 & 27.71/0.799 & 26.58/0.808 \\
        \cline{2-5}
        & GAT+BM3D~\cite{dabov2007image} & 26.16/0.732 & 27.26/0.785 & 26.40/0.795 \\ 
        & N2V~\cite{krull2019noise2void} & 26.25/0.719 & 26.48/0.755 & 24.71/0.741 \\
        & FBI-D~\cite{byun2021fbi} & 25.75/0.683 & 26.42/0.745 & 24.51/0.722 \\
        & NBR2NBR~\cite{huang2021neighbor2neighbor} & 26.88/0.751 & 27.51/0.783 & 26.39/0.800 \\
        & Blind2Unblind~\cite{wang2022blind2unblind} & \underline{27.02}/\underline{0.757} & \underline{27.65}/\underline{0.796} & \underline{26.54}/\underline{0.805} \\
        & Ours & \textbf{27.17}/\textbf{0.766} & \textbf{27.96}/\textbf{0.805} & \textbf{26.96}/\textbf{0.819} \\    
    \bottomrule
  \end{tabular}
  \caption{Grayscale image denoising results.~\textbf{Bold} and \underline{underlined} are the highest and second without supervision.}
  \label{tab:gray}
  \vspace{-0.4cm}
\end{table}

\subsection{Results for Synthetic Denoising}
\noindent\textbf{Grayscale Denoising\;}
The denoising results on synthetic grayscale images are presented in~\Cref{tab:gray}. Three Poisson-Gaussian noise levels are simulated to evaluate the performance of the proposed method under low, medium, and high Poisson-Gaussian noise levels. The results show that our approach outperforms several denoising methods, including the traditional denoising method GAT+BM3D and four self-supervised denoising methods (N2V, NBR2NBR, FBI-D, and Blind2Unblind) for both low and high Poisson-Gaussian noise levels. In addition, our method outperforms two supervised baselines (N2C and N2N) on Set12 and Urban100, with a maximum gain of 0.4 dB, and performs competitively on BSD68. These results highlight the superior generalization of our method on grayscale image denoising compared to supervised baselines. Compared with the recent best Blind2Unblind, our method has a maximum gain of 0.44 dB and a minimum gain of 0.15 dB, demonstrating the necessity of explicit personalized denoising.

\begin{table}[ht]
\scriptsize
  \centering
  \setlength{\abovecaptionskip}{0.1cm} 
  \setlength\tabcolsep{4pt} 
  \begin{tabular}[b]{clccc}
    \toprule
      Noise Type & Method & KODAK & SET14 & BSD300 \\
    \midrule
      \multirow{8}{*}{\shortstack[c]{$\alpha=0.01$\\$\sigma=0.0002$}} 
        & Baseline, N2C~\cite{ronneberger2015u} & 34.67/0.925 & 33.16/0.904 & 33.61/0.929 \\
        & Baseline, N2N~\cite{lehtinen2018noise2noise} & 34.64/0.924 & 33.13/0.904 & 33.59/0.928 \\
        \cline{2-5}
        & GAT+BM3D~\cite{dabov2007image} & 33.63/0.913 & 31.80/0.883 & 32.47/0.909 \\ 
        & N2V~\cite{krull2019noise2void} & 31.68/0.871 & 30.72/0.848 & 29.71/0.844 \\
        & FBI-D~\cite{byun2021fbi} & 31.66/0.871 & 30.66/0.848 & 29.69/0.843 \\
        & NBR2NBR~\cite{huang2021neighbor2neighbor} & \underline{34.10}/\underline{0.918} & \underline{32.69}/\textbf{0.896} & \underline{32.89}/\underline{0.919} \\
        & Blind2Unblind~\cite{wang2022blind2unblind} & 33.88/0.915 & 32.47/0.886 & 32.53/0.913 \\
        & Ours & \textbf{34.23/0.920} & \textbf{32.75/0.896} & \textbf{33.00/0.921} \\      
    \midrule
      \multirow{8}{*}{\shortstack[c]{$\alpha=0.01$\\$\sigma=0.02$}} 
        & Baseline, N2C~\cite{ronneberger2015u} & 34.39/0.920 & 32.93/0.899 & 33.28/0.923 \\
        & Baseline, N2N~\cite{lehtinen2018noise2noise} & 34.36/0.920 & 32.89/0.899 & 33.25/0.923 \\
        \cline{2-5}
        & GAT+BM3D~\cite{dabov2007image} & 33.39/0.909 & 31.58/0.876 & 32.21/0.904 \\ 
        & N2V~\cite{krull2019noise2void} & 31.51/0.867 & 30.56/0.845 & 29.55/0.838 \\
        & FBI-D~\cite{byun2021fbi} & 31.54/0.867 & 30.56/0.846 & 29.58/0.838 \\
        & NBR2NBR~\cite{huang2021neighbor2neighbor} & \underline{33.93}/\underline{0.915} & \underline{32.52}/\underline{0.892} & \underline{32.87}/\underline{0.916} \\
        & Blind2Unblind~\cite{wang2022blind2unblind} & 33.58/0.909 & 32.17/0.883 & 32.31/0.910 \\
        & Ours & \textbf{34.12/0.917} & \textbf{32.60/0.893} & \textbf{32.95/0.917} \\    
    \midrule
      \multirow{8}{*}{\shortstack[c]{$\alpha=0.05$\\$\sigma=0.02$}} 
        & Baseline, N2C~\cite{ronneberger2015u} & 30.80/0.854 & 29.69/0.837 & 29.45/0.843 \\
        & Baseline, N2N~\cite{lehtinen2018noise2noise} & 30.77/0.853 & 29.65/0.836 & 29.43/0.842 \\
        \cline{2-5}
        & GAT+BM3D~\cite{dabov2007image} & 29.19/0.824 & 27.58/0.801 & 27.87/0.799 \\ 
        & N2V~\cite{krull2019noise2void} & 29.03/0.793 & 28.14/0.785 & 27.42/0.759 \\
        & FBI-D~\cite{byun2021fbi} & 29.15/0.800 & 28.29/0.791 & 27.48/0.763 \\
        & NBR2NBR~\cite{huang2021neighbor2neighbor} & 30.49/0.848 & 29.46/0.832 & 29.20/0.837 \\
        & Blind2Unblind~\cite{wang2022blind2unblind} & \underline{30.58}/\underline{0.849} & \underline{29.52}/\underline{0.832} & \underline{29.27}/\underline{0.837} \\
        & Ours & \textbf{30.69/0.850} & \textbf{29.57/0.833} & \textbf{29.33/0.838} \\    
    \bottomrule
  \end{tabular}
  \caption{PSNR(dB)/SSIM on synthetic sRGB datasets.}
  \label{tab:sRGB}
  \vspace{-0.4cm}
\end{table}

\noindent\textbf{sRGB Denoising\;}
The results of synthetic denoising for sRGB images are presented in~\Cref{tab:sRGB}. Our method outperforms GAT+CBM3D and four self-supervised methods (N2V, NBR2NBR, FBI-D, and Blind2Unblind) in both low and high noise levels. However, the gain over Blind2Unblind decreases at high noise levels due to numerous missing details valuable for restoring clean signals. Unlike grayscale denoising in~\Cref{tab:gray}, our method does not outperform supervised baselines in sRGB denoising but achieves competitive performance at high noise levels. Intuitively, cross-channel correlation makes noise removal in sRGB space more challenging than in grayscale images. As the noise level increases, the necessity for clean target supervision gradually decreases. Moreover, we observe that Blind2Unblind with greedy pixel-level objective outperforms NBR2NBR with neighboring approximation at high noise levels, highlighting the potential of lossless denoising for high noise levels. 

\begin{table}[ht]
\scriptsize 
  \centering
  \setlength{\abovecaptionskip}{0.1cm} 
  \setlength\tabcolsep{1.5pt}
  \begin{tabular}[b]{@{}lccccc@{}}
    \toprule
      \multirow{3}{*}[-0.5ex]{Methods} 
      & \multicolumn{3}{c}{SIDD} & \multicolumn{2}{c}{FMD} \\
      \cmidrule(lr){2-4} \cmidrule{5-6}
      & RAW & RAW & sRGB & Confocal & Two-Photon \\
      & Benchmark & Validation & Benchmark & Mice & Mice \\
    \cmidrule{1-1} \cmidrule(lr){2-4} \cmidrule(lr){5-6}
      Baseline, N2C~\cite{ronneberger2015u} & 50.61/0.991 & 51.19/0.991 & 38.08/0.945 & 38.40/0.966 & 34.02/0.925 \\
      Baseline, N2N~\cite{lehtinen2018noise2noise} & 50.62/0.991 & 51.21/0.991 & 38.09/0.945 & 38.37/0.965 & 33.80/0.923 \\
    \cmidrule{1-1} \cmidrule(lr){2-4} \cmidrule(lr){5-6}
      GAT+BM3D~\cite{dabov2007image} & 48.60/0.986 & 48.92/0.986 & 34.64/0.879 & 37.93/0.963 & 33.83/\textbf{0.924} \\
      N2V~\cite{krull2019noise2void} & 48.01/0.983 & 48.55/0.984 & 34.21/0.864 & 37.49/0.960 & 33.38/0.916 \\
      DBSN~\cite{wu2020unpaired} & 49.56/0.987 & 50.13/0.988 & 36.77/0.917 & 30.61/0.730 & 26.24/0.423 \\
      BP-AIDE~\cite{byun2020learning} & 50.45/0.990 & -- & 37.91/0.942 & 38.31/0.963 & 33.89/0.902 \\
      FBI-D~\cite{byun2021fbi} & 50.57/0.990 & -- & 38.07/0.942 & 38.32/0.963 & 33.95/0.908 \\
      NBR2NBR~\cite{huang2021neighbor2neighbor} & 50.47/0.990 & 51.06/0.991 & 37.85/0.942 & 37.07/0.960 & 33.40/\underline{0.921} \\
      Blind2Unblind~\cite{wang2022blind2unblind} & \underline{50.79}/\textbf{0.991} & \underline{51.36}/\textbf{0.992} & \underline{38.11}/\underline{0.944} & \underline{38.44}/\underline{0.964} & \underline{34.03}/0.916 \\
      Ours & \textbf{50.92/0.991} & \textbf{51.50/0.992} & \textbf{38.21/0.945} & \textbf{38.46}/\textbf{0.965} & \textbf{34.11}/0.918 \\
    \bottomrule
  \end{tabular}
  \caption{PSNR(dB)/SSIM on SIDD and FMD dataset.}
  \label{tab:real}
  \vspace{-0.4cm}
\end{table}

\subsection{Results for Real-World Denoising}
\Cref{tab:real} shows the quantitative results for the SIDD benchmark and validation datasets in raw-RGB and sRGB space. The proposed Blind2Sound outperforms all self-supervised methods and supervised baselines, showcasing its strong generalization ability for real-world scenarios with various dynamic Poisson-Gaussian noise levels. Notably, our method achieves a gain of nearly 0.3 dB over FBI-D in raw-RGB space, validating the effectiveness of lossless denoising. Blind2Sound outperforms Blind2Unblind with a gain of 0.1 dB, confirming the importance of explicit personalized modeling. The visual results in~\Cref{fig:sidd} demonstrate that Blind2Sound restores the highest level of texture details and pixel correlation while avoiding bad artifacts. In contrast, Blind2Unblind displays visible clumping shadows in multiple regions due to insufficient denoising. Even supervised baselines show notable artifacts under dynamic noise, further validating the superiority of noise-sensed personalized denoising. The denoising results on FMD dataset indicate that Blind2Sound outperforms self-supervised methods and has competitive performance compared to supervised baselines.

\begin{figure}[h]
\centering
\vspace{-0.2cm}
\setlength{\abovecaptionskip}{0.cm} 
\includegraphics[width=.485\textwidth]{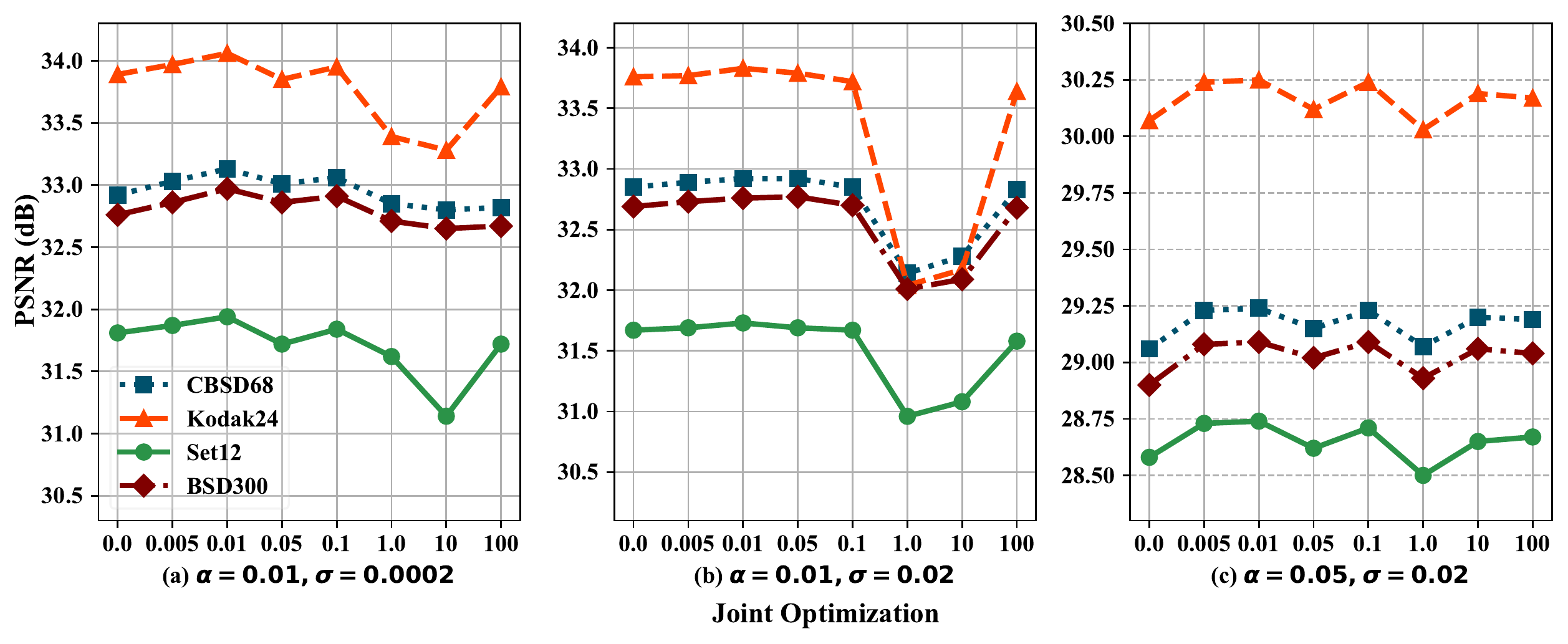}
   \caption{Ablation study on Cramer Gaussian loss with fixed weight $1$ for $L_{arv}$ and horizontal is the weight of $L_{est}$.}
\label{fig:joint}
\vspace{-0.2cm}
\end{figure}

\subsection{Ablation Study}
This section presents ablation studies on various factors, including grain size, Cramer Gaussian loss, training scheme, noise model, re-visible correlation and visible factor. PSNR(dB)/SSIM is evaluated on CBSD68 while training on CBSD432 for 200 epochs.

\noindent\textbf{Ablation study on grain size.}~\Cref{tab:grain} shows that fine-grained sub-block constraints improve the accuracy of the coarse-grained Gaussian loss, enhance robustness, and reduce solution space to provide more accurate noise parameters. However, smaller sub-block sizes results in limited noise context, leading to lower performance.

\noindent\textbf{Ablation study on Cramer Gaussian loss.} As shown in~\Cref{fig:joint}, the denoiser with a weight of $0.01$ performs best compared to $0$ or $100$. Joint optimization can restore more image details and actual noise levels of optimal denoised images may differ from raw noisy images. Hence, Cramer Gaussian loss only serves as a regularization to assist the denoiser in sensing the most suitable noise level.

\noindent\textbf{Ablation study on training scheme.} As shown in~\Cref{tab:opt}, $T+J$ performs better than $T+P$ and $T+F$ at low noise, but they are comparable at high noise. At low noise, $T+P$ and $T+F$ mismatch the actual noise levels of restored images, thus degrading performance. At high noise, three training schemes estimate almost identical noise parameters.

\noindent\textbf{Ablation study on noise model.}~\Cref{tab:mod} shows that $\mathcal{M}_{E}$ performs slightly better than the other noise models at low noise, but much better at high noise. $\mathcal{M}_{O}$ amplifies Poisson noise error in masked branch due to signal dependence, while $\mathcal{M}_{S}$ violates the independence of two branches. 

\noindent\textbf{Ablation study on re-visible correlation.}~\Cref{tab:corr} explores the impact of whether masked and visible branches are independent. The denoiser in the IID setting performs much better than in the non-IID setting. IID decouples the correlation between the blind and visible branches, thus achieving visible denoising without mask suppression.

\noindent\textbf{Ablation study on visible factor.}~\Cref{tab:rev} shows the performance using different visible factors. The degree of visible is not proportional to the performance. Instead, the performance first increases and then decreases as the visible factor increases, reaching a peak when $\lambda_f=11$.

\begin{table}[t]
\scriptsize
  \centering
  \setlength{\abovecaptionskip}{0.1cm} 
  \setlength\tabcolsep{3pt}
  \begin{tabular}[b]{@{}ccccc}
    \toprule
      Grain Size & $(0.01,0.0002)$ & $(0.01,0.02)$ & $(0.05,0.02)$ & $(0.1,0.02)$ \\
    \midrule
      CG & 0.007/0.038 & 0.009/0.034 & 0.039/0.057 & 0.092/0.039 \\
      FG1 & \textbf{0.010}/0.011 & \textbf{0.010/0.021} & 0.046/\textbf{0.021} & 0.089/0.008 \\
      CG+FG1 & \textbf{0.010}/0.012 & \textbf{0.010/0.021} & \textbf{0.048}/0.022 & \textbf{0.093/0.019} \\
      CG+FG2 & 0.009/\textbf{0.008} & \textbf{0.010}/0.022 & 0.043/0.027 & 0.085/0.013 \\
      CG+FG1+FG2 & 0.009/0.009 & \textbf{0.010}/0.022 & 0.046/0.024 & 0.092/0.017 \\
    \bottomrule
  \end{tabular}
  \caption{Ablation study on grain size for BSD68. CG is global coarse grain, FG1 is four sub-blocks each in three-quarter size and FG2 is nine sub-blocks in one-half size.}
  \label{tab:grain}
  \vspace{-0.2cm}
\end{table}

\begin{table}[t]
\scriptsize
  \centering
  \setlength{\abovecaptionskip}{0.1cm} 
  \setlength\tabcolsep{8pt}
  \begin{tabular}[b]{@{}ccccc}
    \toprule
      Training Scheme & $(0.01,0.0002)$ & $(0.01,0.02)$ & $(0.05,0.02)$ \\
    \midrule
      $T+P$ & 32.78/0.914 & 32.69/0.913 & 29.24/0.838 \\
      $T+F$ & 33.02/0.917 & 32.70/0.913 & 29.26/0.839 \\
      $T+J$ & \textbf{33.14/0.920} & \textbf{32.89/0.916} & \textbf{29.29/0.840} \\
    \bottomrule
  \end{tabular}
  \caption{Ablation study on training scheme, $T+P$, $T+F$ and $T+J$ represent training using frozen pre-trained noise estimator, training with fixed true noise or joint training.}
  \label{tab:opt}
  \vspace{-0.2cm}
\end{table}

\begin{table}[t]
\scriptsize
  \centering
  \setlength{\abovecaptionskip}{0.1cm} 
  \setlength\tabcolsep{11pt}
  \begin{tabular}[b]{@{}ccccc}
    \toprule
      Loss Type & $(0.01,0.0002)$ & $(0.01,0.02)$ & $(0.05,0.02)$ \\
    \midrule
      $\mathcal{M}_{O}$ & 33.11/0.919 & 32.84/0.913 & 29.12/0.833 \\
      $\mathcal{M}_{S}$ & 33.10/0.919 & 32.86/0.915 & 29.14/0.835 \\
      $\mathcal{M}_{E}$ & \textbf{33.14/0.920} & \textbf{32.89/0.916} & \textbf{29.29/0.840} \\
    \bottomrule
  \end{tabular}
  \caption{Ablation study on noise model. $\mathcal{M}_{O}$ and $\mathcal{M}_{E}$ denote~\cref{eq:12} and~\cref{eq:13}. $\mathcal{M}_{S}$ means $diag(\alpha\boldsymbol{\mu_y})+\sigma\boldsymbol{\rm{I}}$.}
  \label{tab:mod}
  \vspace{-0.2cm}
\end{table}

\begin{table}[t]
\scriptsize
  \centering
  \setlength{\abovecaptionskip}{0.1cm} 
  \setlength\tabcolsep{11pt}
  \begin{tabular}[b]{@{}ccccc}
    \toprule
      Loss Type & $(0.01,0.0002)$ & $(0.01,0.02)$ & $(0.05,0.02)$ \\
    \midrule
      non-IID & 32.19/0.904 & 32.24/0.905 & 28.95/0.827 \\
      IID & \textbf{33.14/0.920} & \textbf{32.89/0.916} & \textbf{29.29/0.840} \\
    \bottomrule
  \end{tabular}
  \caption{Ablation study on re-visible correlation. non-IID and IID indicate whether the masked and visible branches are independent.}
  \label{tab:corr}
  \vspace{-0.2cm}
\end{table}

\begin{table}[t]
\scriptsize
  \centering
  \setlength{\abovecaptionskip}{0.1cm} 
  \setlength\tabcolsep{5pt}
  \begin{tabular}[b]{@{}ccccccc@{}}
    \toprule
      Noise Type & $\lambda_{f}=3$ & $\lambda_{f}=11$ & $\lambda_{f}=20$ & $\lambda_{f}=40$ \\
    \midrule
      $(0.01,0.0002)$ & 33.05/0.919 & \textbf{33.14/0.920} & 33.11/\textbf{0.920} & 33.06/0.919 \\
      $(0.01, 0.02)$ & 32.75/0.914 & \textbf{32.89/0.916} & 32.84/0.915 & 32.76/0.914 \\
      $(0.05, 0.02)$ & 29.24/0.838 & \textbf{29.29/0.840} & 29.27/0.839 & 29.23/0.837  \\
    \bottomrule
  \end{tabular}
  \caption{Ablation study on visible factor. Note that the initial value $\lambda_{s}=3$ and $\lambda_f$ is the final value.}
  \label{tab:rev}
  \vspace{-0.2cm}
\end{table}

\section{Conclusion}
We propose Blind2Sound, a self-supervised blind denoising framework that removes Poisson-Gaussian noise without residual noise and adapts to sensed noise levels. Adaptive re-visible loss associates  noise parameters with re-visible transitions to achieve personalized and lossless denoising. Cramer Gaussian loss introduces fine-grained noise knowledge to improve estimation accuracy. The noise estimator is removed during inference. Extensive experiments show that our approach achieves superior performance, especially in real scenes with dynamic noise.


\clearpage

{\small
\bibliographystyle{ieee_fullname}
\bibliography{egbib}
}

\end{document}